\documentclass[10pt,twocolumn,letterpaper]{article}

\usepackage{cvpr}              

\usepackage[dvipsnames]{xcolor}

\definecolor{cvprblue}{rgb}{0.21,0.49,0.74}
\usepackage[pagebackref,breaklinks,colorlinks,citecolor=cvprblue]{hyperref}
\usepackage{xcolor}
\usepackage{color}
\usepackage{appendix}
\usepackage{pifont}
\usepackage{graphicx}
\usepackage{multirow}
\usepackage{booktabs}
\usepackage{subcaption} 
\newcommand{\cross}{\ding{55}}%
\renewcommand{\checkmark}{\ding{51}}%
\newcommand{\pparagraph}[1]{\medskip\noindent\textbf{#1}}

\def\paperID{6} 
\def\confName{EarthVision}
\def\confYear{2024}

\title{Exploring Robust Features for Few-Shot Object Detection in Satellite Imagery}

\author{Xavier Bou\textsuperscript{1} \and Gabriele Facciolo\textsuperscript{1} \and Rafael Grompone von Gioi\textsuperscript{1} \and Jean-Michel Morel\textsuperscript{2} \and Thibaud Ehret\textsuperscript{1}\\
\textsuperscript{1}Université Paris-Saclay, CNRS, ENS Paris-Saclay, Centre Borelli, France \\
\textsuperscript{2}City University of Hong Kong, Department of Mathematics, Kowloon, Hong Kong \\
{\tt\small xavier.bou\_hernandez@ens-paris-saclay.fr}
}

\begin{document}
\maketitle
\begin{abstract}
The goal of this paper is to perform object detection in satellite imagery with only a few examples, thus enabling users to specify any object class with minimal annotation. To this end, we explore recent methods and ideas from open-vocabulary detection for the remote sensing domain. We develop a few-shot object detector based on a traditional two-stage architecture, where the classification block is replaced by a prototype-based classifier. A large-scale pre-trained model is used to build class-reference embeddings or prototypes, which are compared to region proposal contents for label prediction. In addition, we propose to fine-tune prototypes on available training images to boost performance and learn differences between similar classes, such as aircraft types. We perform extensive evaluations on two remote sensing datasets containing challenging and rare objects. Moreover, we study the performance of both visual and image-text features, namely DINOv2 and CLIP, including two CLIP models specifically tailored for remote sensing applications. Results indicate that visual features are largely superior to vision-language models, as the latter lack the necessary domain-specific vocabulary. Lastly, the developed detector outperforms fully supervised and few-shot methods evaluated on the SIMD and DIOR datasets, despite minimal training parameters.
\end{abstract}
    
\section{Introduction}
\label{sec:intro}

\begin{figure}[t]
    \centering
    \includegraphics[width=1\linewidth]{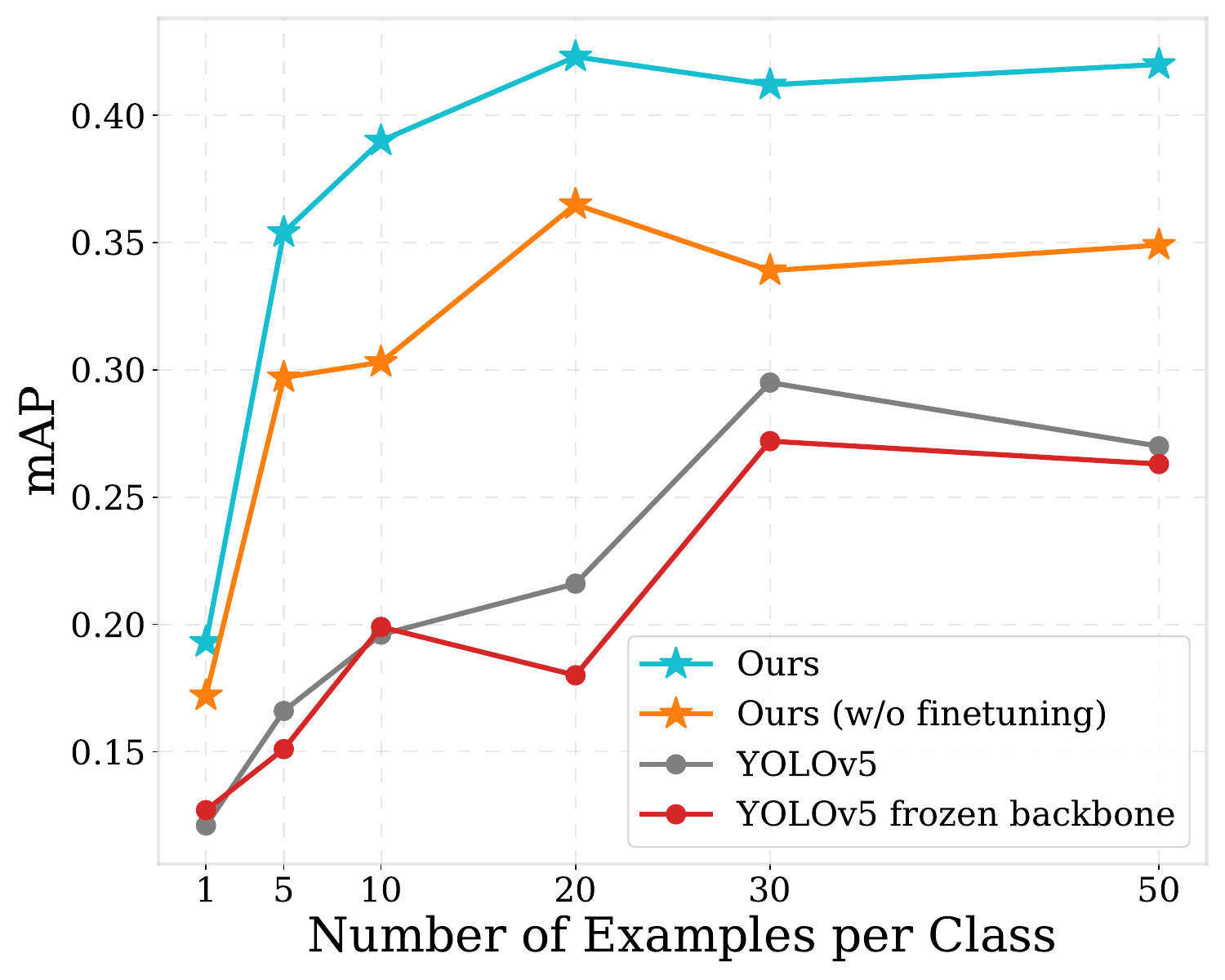}
    \caption{Performance (mAP) of the proposed detector with DINOv2 features on the SIMD dataset, compared to YOLOv5 for different amounts of available examples per class. Robust visual features largely outperform state-of-the-art supervised methods when annotated data is limited.}
    \label{fig:teaser}
\end{figure}

Object detection in remote sensing data is a crucial problem for earth observation applications such as intelligent monitoring, urban planning and precision agriculture~\cite{LI2020296}. It consists of locating and assigning labels to objects of interest in an image. In recent years, fully supervised object detection has shown impressive performances with methods like YOLO~\cite{yolo}, Faster RCNN~\cite{fasterrcnn} or Mask-RCNN~\cite{maskrcnn}. However, these methods require large amounts of annotated data, which poses an issue in the context of remote sensing, where available annotated data are scarce.

More recently, the emergence of large-scale vision-language models (VLMs) introduced the problem of Open-Vocabulary Detection (OVD), which aims to detect objects beyond the set of training classes. OVD methods try to exploit the knowledge of these pre-trained VLMs to perform detection conditioned by their image and/or text embeddings~\cite{vild, regionclip, ovdetr, owlvit}. Conveniently, this leads to a drastic reduction of the annotation cost and enables the use of text prompts, for zero-shot, or a reference image, for one-shot, to specify novel target classes.

Despite showing impressive capabilities to detect rare object classes that are uncommon amongst popular datasets, the amount of aerial or satellite image examples used during training is minimal. Hence, due to the large gap between natural images and optical remote sensing data, the performance of proposed OVD methods on the latter is quite poor. In addition, a known issue in OVD methods is the need for prompt engineering to find the most suitable wording for desired object classes. It has been indeed shown that a slight word change can negatively or positively affect detection performance~\cite{learning_tokens, learning_tokens_old}. For this reason, some works have proposed approaches to automatically find the most suitable adjectives~\cite{learning_adjectives}, text tokens~\cite{learning_tokens} or embeddings~\cite{multimodal_classifiers_zisserman} to improve classification using a few examples, framing the problem as Few-Shot Object Detection (FSOD). Furthermore, DE-ViT~\cite{learning_tokens} recently introduced the idea that CLIP text embeddings are not discriminative enough, proposing a framework using purely visual DINOv2 features~\cite{dinov2}.

In this work, we revisit the ideas of OVD and FSOD based on large-scale pre-trained models and explore their capabilities for FSOD in the remote sensing domain. To this end, we re-purpose a two-stage object detector for FSOD, where the classification step is replaced by an OVD-inspired classifier that uses feature embeddings as reference classes. Furthermore, the limited class examples are used to fine-tune the class reference embeddings in automatic prompt-engineering mode, learning the difference between target objects and background classes.

We compare the performance of our detector with other FSOD and fully supervised methods. In addition, we explore several robust features for both visual and vision-language models, including RemoteCLIP~\cite{remoteclip} and GeoRSCLIP~\cite{georsclip}, which are specifically tailored for remote sensing applications. Our results indicate that visual features are more suitable for remote sensing detection, as image-text approaches seem to be limited by the granularity of their image captions. Furthermore, DINOv2 representations show an impressive ability to discriminate similar types of rare classes, such as types of aircraft or vehicles. The proposed detector outperforms all other evaluated methods on the SIMD and DIOR datasets. Figure~\ref{fig:teaser} illustrates the detection advantage of the proposed framework over a fully supervised approach for minimal annotations.
\section{Related works}
\label{sec:related_works}
\pparagraph{Closed-vocabulary object detection} is a traditional problem in image understanding that attempts not only to identify objects in images but to precisely estimate their location as well~\cite{obj_det_review_1}. In recent years, advances in deep neural networks have led to great success in the field, and such approaches have consolidated as the state of the art over classical algorithms~\cite{obj_det_review_2}. Current-day object detectors can generally be divided into two-stage and one-stage detectors. Two-stage object detectors proceed in two steps. First, a region proposal network (RPN) extracts region-of-interest (RoI) features and generates class-agnostic bounding box proposals, which are later classified by a sub-network~\cite{rcnn, fasterrcnn, maskrcnn}. On the other hand, one-stage detectors avoid the time-consuming proposal generation and work directly over a dense sampling of locations or anchors~\cite{yolo, retinanet, mobilenet}. While one-stage detectors achieve faster inference, this can come at the cost of performance. Despite recent progress, closed-vocabulary object detection methods are limited to detecting only classes seen during training, requiring significant annotation and training effort to extend them to new categories.

\pparagraph{Few-Shot Object Detection} aims to detect objects in images with limited annotated data to handle the absence of objects in common large-scale datasets. The general underlying idea is training a detector on a set of base classes with a large amount of annotated bounding boxes, and then adapting the classification step to perform for new, unseen classes using only a few examples~\cite{FSOD_review}. Kang \textit{et al.}~\cite{fsrw} proposed a feature re-weighting module (FSRW) that quickly learns to use general meta-features to detect novel classes. Snell \textit{et al.}~\cite{prototypical_networks} proposed a simple approach to few-shot classification called prototypical networks, where each class is represented by a mean vector of the embedded support points belonging to its class. We build on this idea in our study.

\pparagraph{Open-Vocabulary Detection} tries to detect objects beyond the set of categories seen during training, i.e. reducing the need for re-training. It has recently gathered attention in the literature, propelled by the development of large-scale vision-language models such as CLIP~\cite{clip}, which learns relationships from both image and text with a shared embedding space. OVD was introduced by Zareian \textit{et al.}~\cite{ovd_first}, who provided a framework to learn both weakly supervised and zero-shot capabilities. Then, Vild~\cite{vild} proposed to use CLIP to match text embeddings of novel classes to corresponding representations in image crops, so that unseen objects are supported at inference. RegionCLIP~\cite{regionclip} generates location pseudo-labels from image-caption pairs, which are then used to align region-text pairs in the feature space via self-supervised learning. OWL-ViT~\cite{owlvit, owlvit_v2} scaled these methods with a simple transformer-based architecture in two steps. First, a vision-language model is trained, which is then fine-tuned to perform object detection conditioned by CLIP embeddings.   As image and text share a common embedding representation, OWL-ViT can perform detection conditioned by an image instead of text, thus enabling one-shot detection. More recently, Kaul \textit{et al.}~\cite{multimodal_classifiers_zisserman} developed a framework to build multi-modal OVD classifiers. They pre-train a class-agnostic RPN and train an aggregator network that utilizes CLIP embeddings from a few examples to generate a classifier vector. DE-ViT~\cite{devit}, on the other hand, diverges from reliance on large vision-language models, opting for purely visual DINOv2~\cite{dinov2} features instead. Class reference vector prototypes are built for each class using a few examples, which are then used for the classification of region proposals. A distinctive work by Parisot \textit{et al.}~\cite{learning_tokens} illustrates the significant impact of prompt engineering on OVD, and derives an approach to fine-tune learnable text tokens, thus reaching the most suitable category prompt for a set of examples. While some recent works require a few examples and thus diverge from OVD towards few-shot detection, they reveal great potential for the detection of rare objects or concepts. Moreover, commonly used large-scale vision-language and vision models are tailored for natural images, as the number of remote sensing data seen during training is minimal. Our work evaluates their performance on remote-sensing data.

\pparagraph{Object Detection in Remote Sensing} is a fundamental problem in the field of aerial image analysis, assuming an important role in a number of applications due to the increasing availability of satellite images~\cite{obj_det_survey_remote_sensing}. The success of deep learning-based object detection in natural images inspired the earth observation community to drive efforts to object detection in optical remote sensing data. As a consequence, several benchmarks were proposed~\cite{dota, vedai, LI2020296, simd}, and various works developed tailored approaches for the characteristics of satellite imagery, e.g. rotation-invariance or small object detection~\cite{obj_rs_1,obj_rs_2,obj_rs_3}. Nevertheless, most remote sensing datasets present shortcomings, such as a low number of categories, class imbalance or insufficient image diversity and variation~\cite{LI2020296}. These constraints obstruct the progress of traditional deep learning-based object detectors, as they require large amounts of well-annotated, curated data. Additionally, some works have explored few-shot methods in this domain. Deng \textit{et al.}~\cite{FSD_RS_1} introduces a reweighting module that re-calibrates feature maps from a set of labelled support images on a YOLO architecture. Zhang \textit{et al.}~\cite{FSD_RS_2} proposes a few-shot approach focusing on avoiding catastrophic forgetting of the base classes. Wolf \textit{et al.}~\cite{wolf2021double} designed a double-head architecture to prevent knowledge loss of base classes, paired with a sampling and pre-processing strategy to better exploit base class annotations. Cheng \textit{et al.}~\cite{pcnn} introduced a prototypical approach based on ResNet101~\cite{resnet} features that adapts a two-stage architecture to detect and classify objects based on support images. More recently, Lu \textit{et al.}~\cite{text_modal} proposed to fuse visual features with text description features for each category, reducing the classification confusion of novel classes.

Nevertheless, few works attempt few-shot detection beyond common, general classes, e.g. \textit{airplane} or \textit{vehicle}. Furthermore, despite some works have adapted image-text pre-training strategies to the remote sensing domain~\cite{remoteclip, georsclip}, no attempts have been made to use these tools and ideas for detection in optical remote sensing data. Consequently, our study on robust representations for few-shot detection in optical remote sensing, akin to recent works on OVD and FSOD for natural images, is relevant and of interest to the community.
\section{Methodology}

\begin{figure}[t]
    \centering
    \includegraphics[width=1\linewidth,trim={0.75cm 6cm 14.1cm 0.25cm},clip]{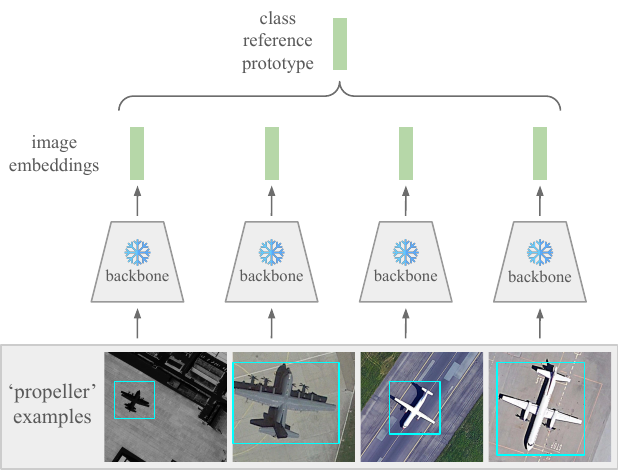}
    \caption{Building a class reference prototype for the aircraft category \textit{propeller} with four examples. The frozen pre-trained backbone is used to extract image representations. Then, patches overlapping box annotations are averaged into one single vector. Lastly, all four embeddings are combined into a reference vector via averaging and normalization.}
    \label{fig:building_prototypes}
\end{figure}

\begin{figure*}[t]
    \centering
    \includegraphics[width=1\linewidth,trim={0.5cm 7.5cm 7.6cm 0.45cm},clip]{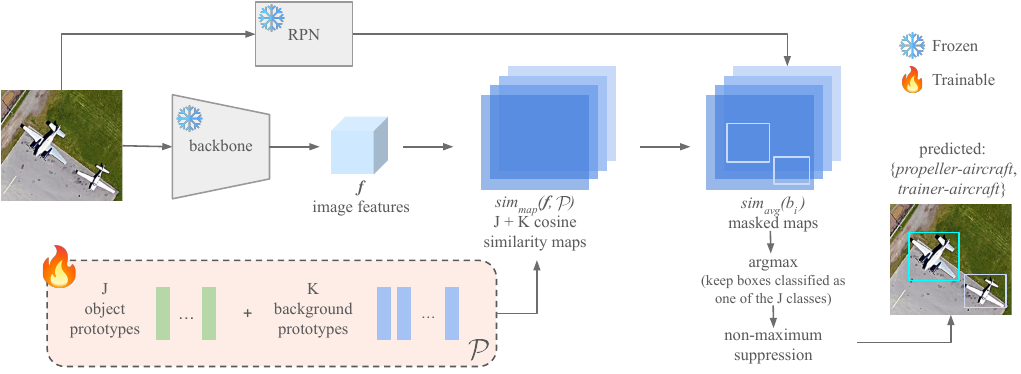}
    \caption{General diagram of our detector. An input image is fed to the RPN to generate region proposals, as well as to the backbone to extract high-level representations. Then, cosine similarity maps are generated using the features and pre-computed prototypes. For each region proposal, the mean average similarity with each prototype is computed, and the proposal is then classified as the most similar prototype class. Lastly, we discard boxes classified as a background prototype and apply non-maximum suppression.}
    \label{fig:detector_inference}
\end{figure*}
In this section, we first frame the problem 
we are addressing. Subsequently, a comprehensive breakdown of our detector is provided, including the architecture overview, how class reference prototypes are built, the classification step, and how prototypes are fine-tuned to improve their classification capabilities.

\pparagraph{Problem setup.}
We consider the problem of detecting objects in optical remote sensing data with limited annotations. We assume to have two remote sensing datasets: a large dataset containing general, common object classes $\mathcal{D}^{train}$, and a dataset containing a very limited number of training instances $\mathcal{D}^{test}$. The objects in $\mathcal{D}^{test}$ can be grouped into base classes $c_{base}$ and novel classes $c_{novel}$, where the first correspond to those contained in $\mathcal{D}^{train}$, and the latter belong to object classes that have never been seen by the model before. Our goal is to detect $c_{novel}$ objects from the limited examples available.

The detector takes an image $I\in{\rm I\!R}^{3\times H\times W}$ as input and predicts the location and class of present objects. For the $i^{th}$ predicted object in an image, bounding box coordinates $b_i\in{\rm I\!R}^4$ and a class prediction $\hat{c_i}\in{\mathcal{C}}$ are generated, where $\mathcal{C}$ is a set of classes such that $\mathcal{C} = c_{\text{base}} \cup c_{\text{novel}}$.

\pparagraph{Architecture overview.}
Following the approach of DE-ViT~\cite{devit} and Kaul \textit{et al.}~\cite{multimodal_classifiers_zisserman}, we opt for a standard two-stage architecture and use a Faster-RCNN~\cite{fasterrcnn} as region proposal network.
For classification, we extract robust image features using a pre-trained backbone, which consists of a Transformer encoder architecture~\cite{vit} that transforms an image $I\in{\rm I\!R}^{3\times H\times W}$ into a high-dimensional feature representation $f$:
\begin{equation}\label{eq:backbone_encode}
     Backbone(I) = f\in{\rm I\!R}^{\frac{H}{p_{s}}\times \frac{W}{p_{s}}\times D},
\end{equation}
where $H$ and $W$ are the image height and width, and $p_{s}$ and $D$ the transformer patch size and feature dimensionality, respectively. Class reference prototypes are then built to represent object categories, and region proposals are classified via cosine similarity between prototypes and image features. This process is detailed in the following sections.

\subsection{Building prototypes}
\label{sec:method_building_prototypes}

\pparagraph{Object prototypes. }
We want to build a set of reference prototype embeddings $\mathcal{P}=\{p_1, p_2, 
\ldots, p_J\}$ that represent target classes $\mathcal{C}=\{c_1, c_2, \ldots, c_J\}$ and can be used to classify region proposals. Hence, for an object category $c_j$ containing $N_j$ examples, its representative prototype $p_j$ is computed as
\begin{equation}\label{eq:built_ptototype}
     p_j = \frac{\hat{p}_j}{\|\hat{p}_j\|_2}, \; \; \text{where} \; \; \hat{p}_j = \frac{\Sigma_{n=1}^{N_j} \sum_{(l,h) \in b_n} (f_n)_{l,h}}{NXY},
\end{equation}
$b_n$ is the bounding box annotation of the $n^{th}$ object, $(f_n)_{l,h}$ is the feature representation of image $n$ at position $(l,h)$, and $X$ and $Y$ are and the width and height of the object ground truth bounding box $b_{n}$, respectively. Figure~\ref{fig:building_prototypes} depicts the process of building a prototype for a specific class.

\pparagraph{Background prototypes } are additionally built to reduce the number of false alarms, addressing the case of invalid region proposals. While the background appearance of natural images is highly variable and often unrelated to object categories, satellite imagery contains a finite number of backgrounds, i.e. the different earth land cover types (water, pavement, urban, forest, etc.). For this reason, we propose to generate $K$ background prototypes using object-free areas in available images. To this end, we extract image representations of all available images and generate a number of crops per image that do not overlap with any labelled instance. Subsequently, we cluster these representations into K clusters using K-Means, and generate a background prototype per cluster by averaging the embeddings in each cluster. Overall, our detector uses $J$ object prototypes and $K$ background prototypes $\mathcal{P}=\{p_1, p_2, ..., p_{J+K}\}$, which are initialized offline.

\subsection{Classification}
\label{sec:method_classification}
During inference, input image $I$ is fed to the detector and a set of region proposals are generated by the RPN. Classification is subsequently performed as follows: we first extract image features $f$ by applying the pre-trained backbone to image $I$ and upsample it to input image resolution. Then, a similarity map is generated by computing the cosine similarity between extracted features and pre-computed prototypes $\mathcal{P}$:
\begin{equation}\label{eq:cosim_maps}
sim_{map}(f, \mathcal{P})= {f \cdot \mathcal{P} \over \|f\| \|\mathcal{P}\|} \in{\rm I\!R}^{H\times W \times (J+K)}.
\end{equation}
Afterward, we extract the $sim_{map}$ similarity values inside each bounding box proposal $b_i\in{\rm I\!R}^4$, and compute the average bounding box similarity for every prototype as follows:
\begin{equation}\label{eq:box_avg_sim}
sim_{avg}(b_i) = \frac{\sum_{(l,h) \in b_i} sim_{map}(f, \mathcal{P})_{l,h}}{XY}\in{\rm I\!R}^{J+K},
\end{equation}
where $ sim_{map}(f, \mathcal{P})$ is the cosine similarity map and X and Y are the width and height of $b_i$, respectively.
Then, $b_i$ is classified as the class of higher $sim_{avg}(b_i)$ according to its prototype. Boxes classified as a background prototype are discarded. Lastly, non-maximum suppression is applied to avoid detecting the same object multiple times.

\subsection{Fine-tuning prototypes}
\label{sec:method_finetuning_prototypes}
As detailed in Section~\ref{sec:method_building_prototypes}, prototypes are built by averaging available object representations into a single reference vector. Nevertheless, bounding boxes include not only the object of interest but a portion of the background as well, introducing undesirable information into class prototypes. Hence, averaging their features might not be the most appropriate approach to classify such cases. Inspired by Parisot \textit{et al.}~\cite{learning_tokens}, which uses learnable word embeddings to find the best suited class names for unseen categories given a VLM, we derive a fine-tuning approach to improve the discriminative capabilities of our prototypes. Similarly, we propose to fine-tune the pre-computed prototypes $\mathcal{P}$ to learn better representations for each class, given available images.

To this end, we optimize prototypes to classify available ground truth boxes between the set of object and background classes. In addition, we randomly sample image crops that do not intersect with object ground truth bounding boxes, and use them as negative examples that need to be classified as background prototypes. For each negative example, we define its ground truth as the class with the most similar prototype amongst the set of background prototypes. More formally, the classifier of our detector predicts a class $\hat{c}_i$ given a region proposal $b_i$, an image feature representation $f$ and set of prototypes $\mathcal{P}$:
\begin{equation}\label{eq:classifier_pred}
Classifier(b_i, f, \mathcal{P}) = \hat{c}_i.
\end{equation}
Thus, learning a set of prototypes $\hat{\mathcal{P}}$ that optimizes the cross-entropy loss objective function over the annotated bounding boxes, given their ground truths. The weights of the backbone are kept frozen at all time.


\section{Experiments}
\begin{table*}[t]
\centering
\begin{tabular}{lccccccc}
\hline
\textbf{Method}         & \textbf{Backbone} & \multicolumn{2}{c}{\textbf{5-shot}} & \multicolumn{2}{c}{\textbf{10-shot}} & \multicolumn{2}{c}{\textbf{30-shot}}   \\ 
                    \cmidrule[0.4pt](lr{0.125em}){3-4}%
                    \cmidrule[0.4pt](lr{0.125em}){5-6}%
                    \cmidrule[0.4pt](lr{0.125em}){7-8}%
                      &  & \textbf{SIMD}  & \textbf{DIOR} & \textbf{SIMD}  & \textbf{DIOR}    & \textbf{SIMD}  & \textbf{DIOR}   \\ 
                        \cmidrule[0.4pt](lr{0.125em}){3-3}%
                        \cmidrule[0.4pt](lr{0.125em}){4-4}%
                        \cmidrule[0.4pt](lr{0.125em}){5-5}%
                        \cmidrule[0.4pt](lr{0.125em}){6-6}%
                        \cmidrule[0.4pt](lr{0.125em}){7-7}%
                        \cmidrule[0.4pt](lr{0.125em}){8-8}%
YOLO       & YOLOv5    & 16.60 & 4.23 &  19.57 & 10.28 & 29.48 & 16.99    \\
YOLO & YOLOv5 (frozen) & 15.05 & 5.70   & 19.94 & 9.42 & 27.18 & 14.90       \\
FSRW~\cite{fsrw} & DarkNet-19  & 11.04 & 10.20  & 13.70 & 15.06   & 23.77 & 25.79 \\
DE-ViT~\cite{devit} & ViT-L/14 & 20.43 & 9.12 & 20.44 & 8.95    & 20.06 & 9.33   \\
Ours & ViT-L/14 & \textbf{35.44} & 9.56 & \textbf{38.99} & 12.51 & 41.21 & 12.60 \\
Ours + FSRW & ViT-L/14 & 29.14 & \textbf{15.06} & 38.61 & \textbf{18.77} & \textbf{41.40} & \textbf{26.46} \\
\hline
\end{tabular}
\caption{Results (mAP50) on DIOR and SIMD benchmarks for 5-shot, 10-shot and 30-shot detection. Several representative object detection methods are evaluated, including fully supervised and FSOD approaches. In the last row, we use the FSRW approach as RPN, and we re-classify each bounding box using the learned prototypes.}
\label{table:main_table_results}
\end{table*}

\pparagraph{Implementation details.} We use the visual model DINOv2~\cite{dinov2} and the vision-language model CLIP~\cite{clip} as pre-trained backbones. We evaluate different versions for the latter, including two remote sensing-tailored models. While fine-tuning prototypes, we use $K=200$ background prototypes and explore different numbers of negative region proposals per image. We apply several spatial augmentations to input images, consisting in random horizontal and vertical flips, random rotations, color jitter, padding and random resized crops of size $602\times 602$. Detailed information on the training setup and hyper-parameters is provided in the supplementary material.

\pparagraph{Experimental setup.} We select the DOTA~\cite{dota} dataset as $\mathcal{D}^{train}$, which contains 2,806 large images with 403,318 annotated instances of 16 general classes, such as \textit{plane}, \textit{ship}, \textit{small vehicle} or \textit{storage tank}. We pre-process the dataset to obtain images of size $800 \times 800$ with an overlap of 200 pixels. Then, a Faster-RCNN model is trained on the entire train set, and the resulting RPN is extracted to serve as our region proposal network.
We consider SIMD~\cite{simd} and DIOR~\cite{LI2020296} as test datasets $\mathcal{D}^{test}$. The SIMD dataset comprises 5,000 images of resolution 1024 × 768 with 45,096 annotated objects, which consist of several aircraft types, e.g. \textit{propeller-aircraft}, \textit{fighter-aircraft} and \textit{airliner}, amongst other more general vehicles, e.g. \textit{car}, \textit{van} or \textit{truck}. Refer to the supplementary material for a detailed description of the base and novel classes of each dataset. The DIOR dataset contains 23,463 images and 192,472 annotations over 20 different classes. While some categories are shared with the DOTA dataset, other significantly different classes are found, such as \textit{chimney}, \textit{express-toll-station}, \textit{airport} or \textit{trainstation}.
To address the few-shot detection performance of novel classes, we generate test subsets for different numbers of examples per class $N = \{5, 10, 30\}$. Class imbalance is highly common in optical remote sensing, as pointed out in Section~\ref{sec:related_works}, and we often observe multiple annotations in a single image. Thus, randomly generating a subset of exactly $N$ instances for all classes can be challenging. For this reason, certain classes in our subsets can contain slightly more or fewer examples than $N$. All data splits will be publicly released, and detailed information can be found in the supplementary material. Lastly, we report the mAP50 scores on novel classes to measure the few-shot detection performance of all evaluated models.

\begin{figure*}
    \centering
    \includegraphics[width=0.245\linewidth,trim={105 50 105 40},clip]{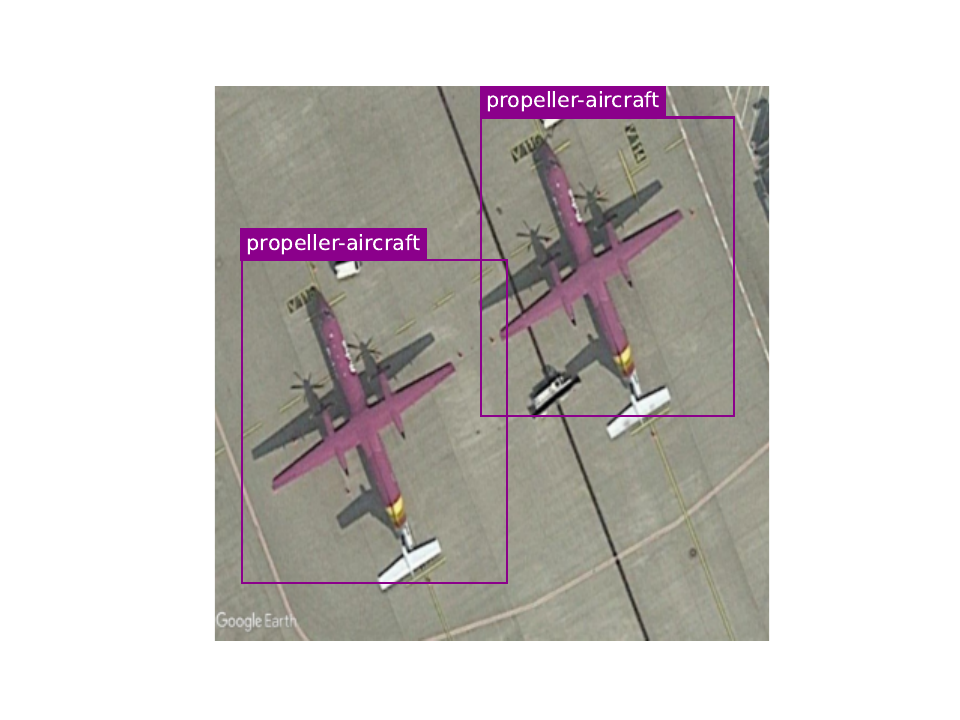}
    \includegraphics[width=0.245\linewidth,trim={105 50 105 40},clip]{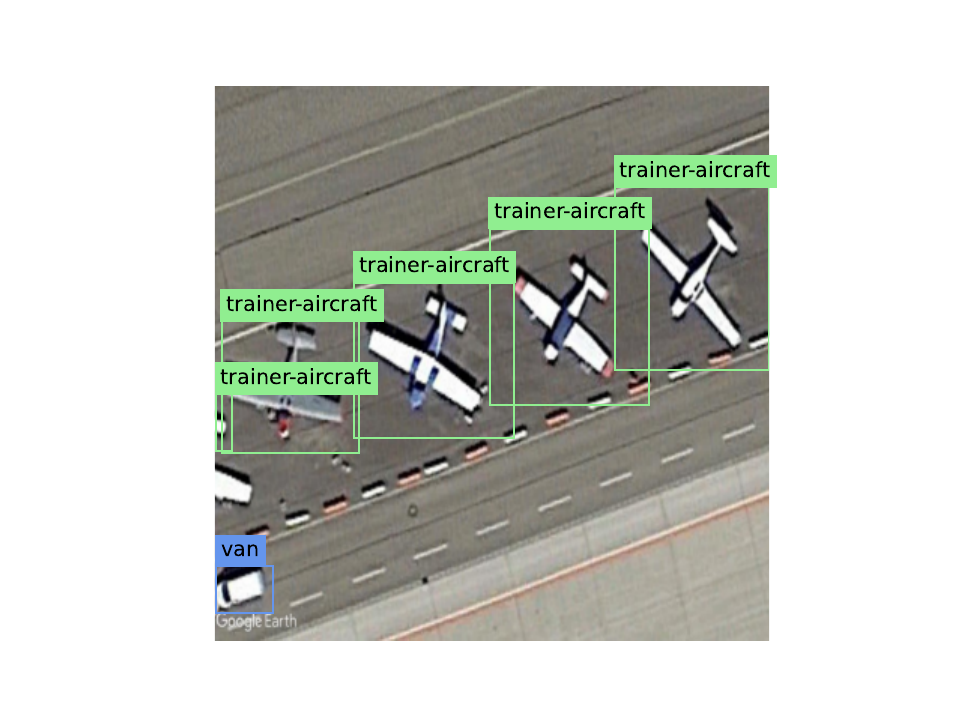}
    \includegraphics[width=0.245\linewidth,trim={105 50 105 40},clip]{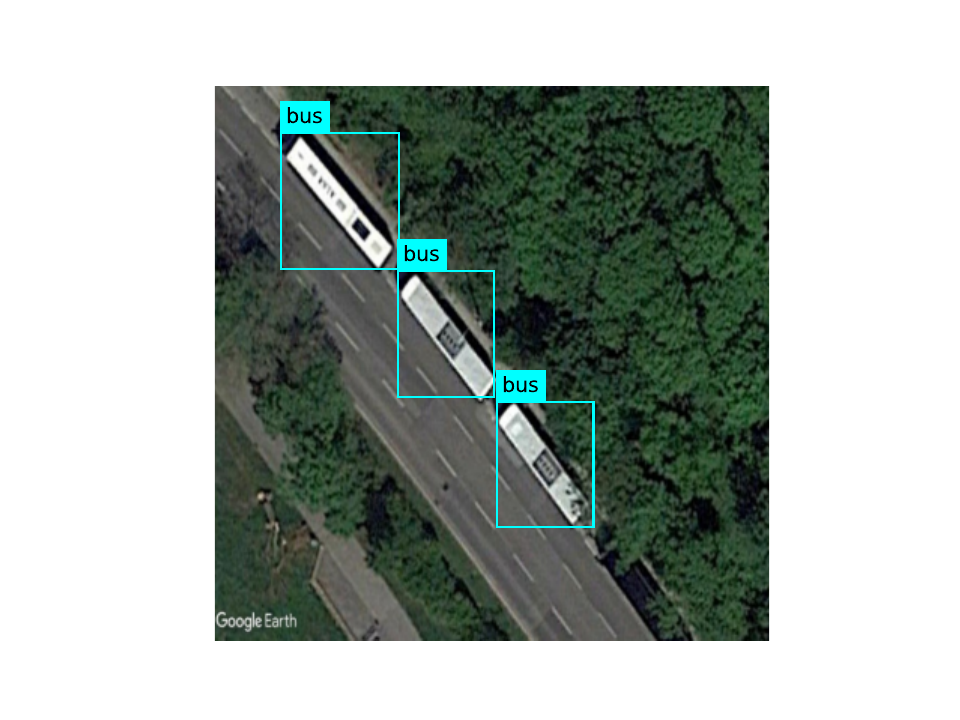}
    \includegraphics[width=0.245\linewidth,trim={105 50 105 40},clip]{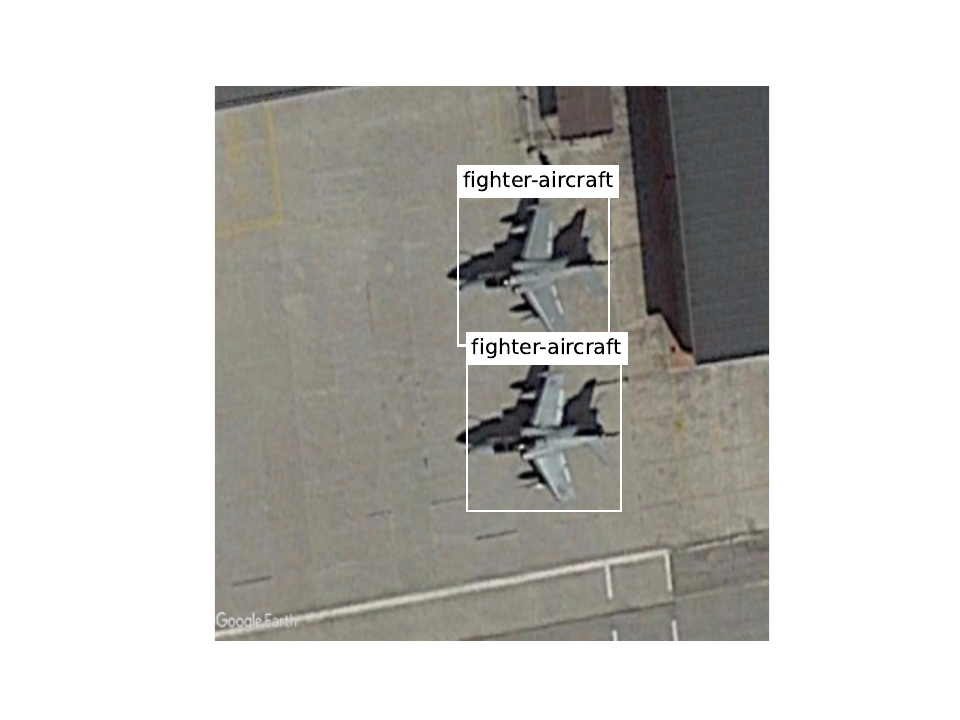}
    \includegraphics[width=0.245\linewidth,trim={105 50 105 40},clip]{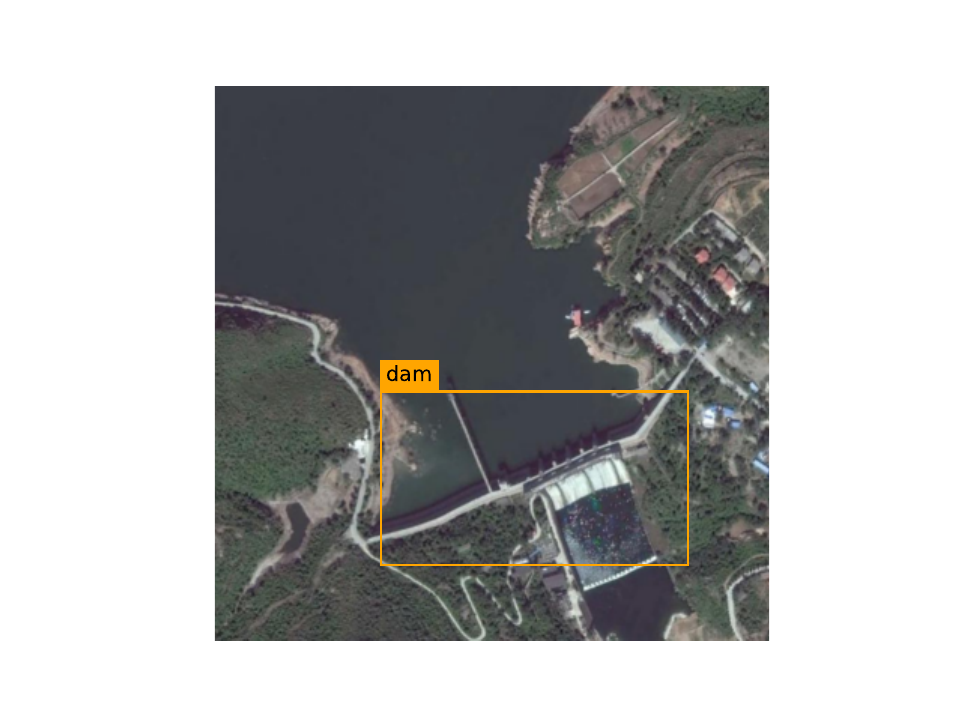}
    \includegraphics[width=0.245\linewidth,trim={105 50 105 40},clip]{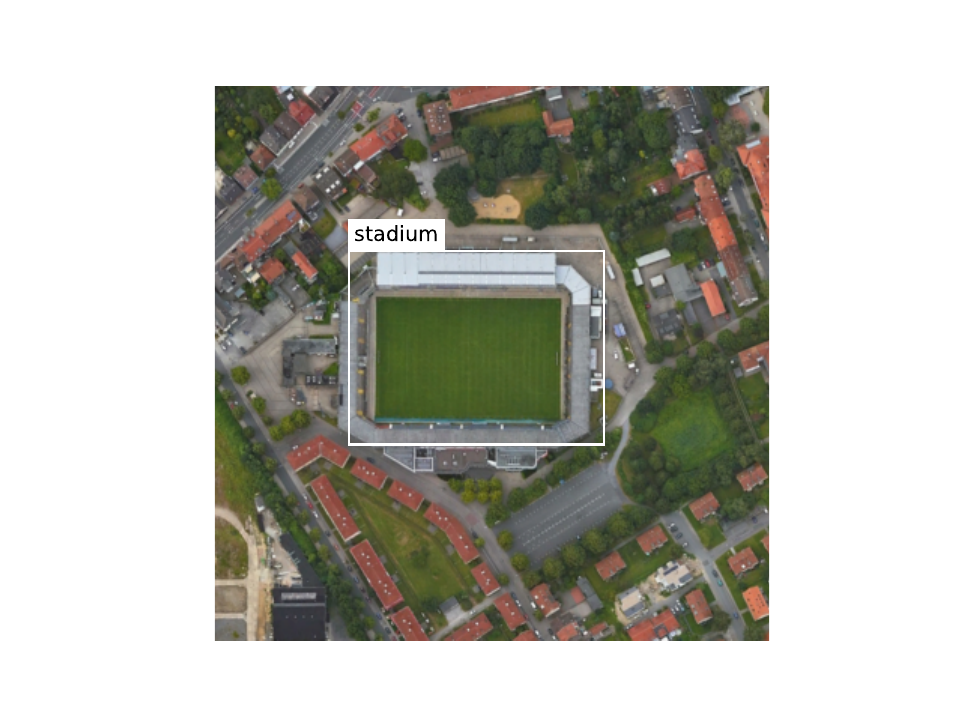}
    \includegraphics[width=0.245\linewidth,trim={105 50 105 40},clip]{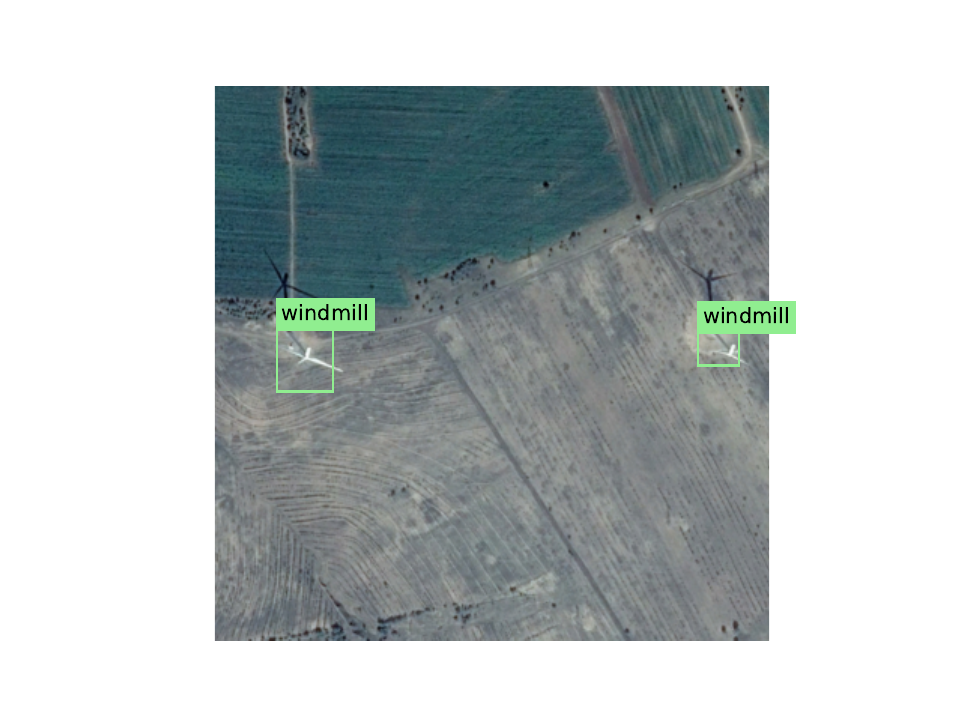}
    \includegraphics[width=0.245\linewidth,trim={105 50 105 40},clip]{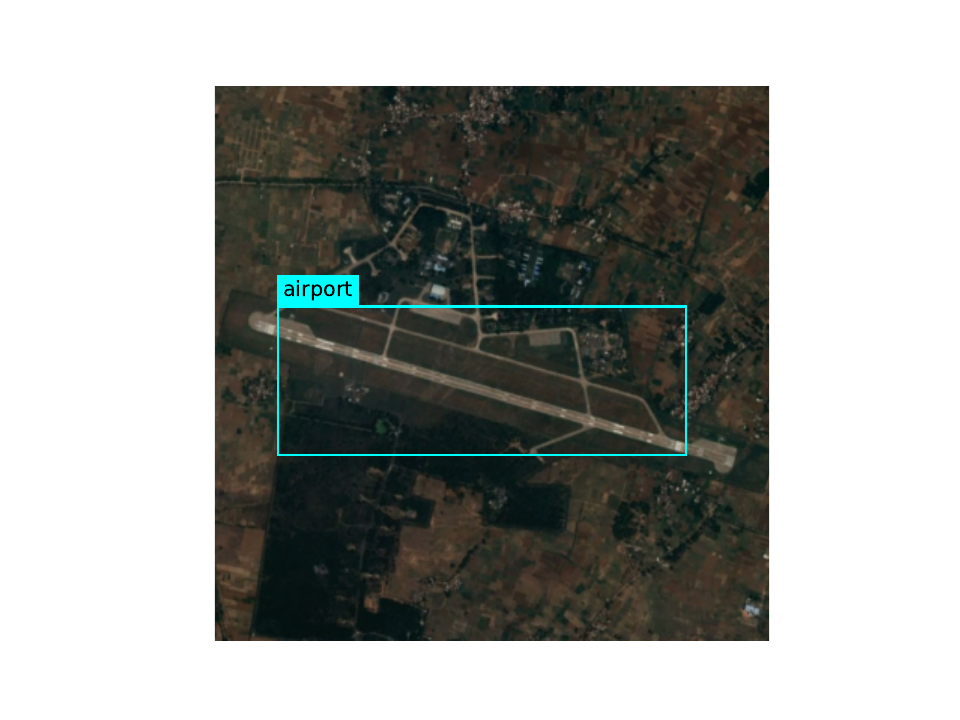}
    \caption{Illustrative qualitative results obtained by the proposed detector. Images on the top row correspond to the SIMD dataset, while images on the bottom belong to the DIOR dataset.}
    \label{fig:qualitative_results}
\end{figure*}

\pparagraph{Results.} We compute the mAP50 results for novel classes on 5-shot, 10-shot and 30-shot of our detector, and benchmark our approach to other methods from the literature. Firstly, we select YOLO as a reference for fully supervised approaches. Hence, we pre-train a YOLOv5 network on the entire DOTA dataset and subsequently fine-tune it using the available few-shot data for each case. Moreover, we consider two relevant FSOD methods, namely the feature reweighting approach (FSRW) introduced by Kang \textit{et al.}~\cite{fsrw} and DE-ViT~\cite{devit}, which uses a prototypical approach with DINOv2 features as well. We observe that for the DIOR dataset, which contains novel classes that are significantly different from the objects in DOTA, the performance of the proposed detector exhibits a decline with respect to the SIMD results. We explore this in detail in Section~\ref{sec:ablation_study}, concluding that the RPN is a limiting block in cases where target objects notoriously differ from the objects in $\mathcal{D}^{train}$. For this reason, we re-use the FSRW model as RPN and re-classify its proposals using our learned prototypes. All results are provided in Table~\ref{table:main_table_results}. As shown, our approach reports large improvements concerning all evaluated methods for the SIMD dataset. It is worth mentioning that while all other methods have previous knowledge of the class \textit{airplane}, they struggle to learn the differences between plane types with minimal examples. Conversely, learning representative DINOv2 prototypes proved to be very discriminative with only a handful of exemplars. It is important to note that our approach uses few training parameters, as only the learnable prototypes are optimized. As for the DIOR dataset, combining the prototypical approach with FSFR yields the best results. This illustrates both the potential of DINOv2 prototypes to classify region proposals and the limitations of pre-training the RPN on the DOTA dataset. As FSRW re-weights pre-trained RoI features of a one-stage detector, the region proposals improve with respect to the base training, resulting in better-suited region proposals for classes that largely diverge from the categories in $\mathcal{D}^{train}$. Qualitative results for the SIMD and DIOR datasets are illustrated in Figure~\ref{fig:qualitative_results}.


\subsection{Ablation study}
\label{sec:ablation_study}
While the best results obtained by the proposed detector are reported in Table~\ref{table:main_table_results}, we conduct in-depth ablation studies on the key components of our approach. Hence, we expand on (1) the choice of pre-trained backbone, (2) the classification abilities of learned DINOv2 prototypes, 
and (3) the impact of fine-tuning prototypes as described in Section~\ref{sec:method_finetuning_prototypes}.

\pparagraph{Visual vs. vision-language features.} Despite VLM having become popular for OVD and FSOD, a question arises when selecting a pre-trained backbone: are vision-language features superior to purely visual features? We aim to shed some light on this issue by comparing DINOv2 and CLIP representations. Furthermore, we add to our analysis two CLIP-based VLMs developed for the remote sensing domain: RemoteCLIP~\cite{remoteclip} and GeoRSCLIP~\cite{georsclip}. To this end, we evaluate our detector on the SIMD dataset for different backbones and $N=10$ examples per class. Results are shown in Table~\ref{table:backbone_ablation}, which displays mAP50 scores for both base and novel classes. Average prototypes, as described in Section~\ref{sec:method_building_prototypes}, are used on the top part, while fine-tuned prototypes, as shown in Section~\ref{sec:method_finetuning_prototypes}, are reported on the bottom. Visual features show a clear advantage over vision-language features in novel classes, even for   RemoteCLIP and GeoRSCLIP. They also report strong results on base classes, despite RemoteCLIP and GeoRSCLIP achieving higher results. This can be explained by two factors: On one hand, base classes correspond to those contained in DOTA, consisting of very general and common objects in remote sensing datasets. We argue that remote sensing datasets are highly overfitted by such classes, e.g. \textit{plane}, \textit{small vehicle}, \textit{ship}, \textit{storage tank}, \textit{tennis court}, etc. Thus, both RemoteCLIP and GeoRSCLIP have repeatedly seen the concepts related to base classes. On the other hand, novel classes consist in exceptionally rare object categories, including \textit{propeller-aircraft}, \textit{airliner}, \textit{charted-aircraft}, or \textit{stair-truck}. Hence, such fine-grained vocabulary is not known by CLIP models, which rely on image captions to learn image-text representations. These captions often lack the ability to describe all elements in the image, since a single satellite image can contain numerous instances and concepts. Thus, we argue that VLMs are limited by the granularity of image descriptions, which restricts their capabilities for FSOD on fine-grained, rare categories.

\begin{table}[t]
\centering
\resizebox{\columnwidth}{!}{
\begin{tabular}{ccccc}
\hline
       \textbf{Backbone}          & \textbf{Fine-tuned} & \textbf{Architecture} & $\textbf{c}_{novel}$ & $\textbf{c}_{base}$ \\ \hline
 CLIP       & & ViT-B/32      & 0.113         & 0.201        \\
 CLIP       & & ViT-L/14      & 0.236         & 0.306        \\
 GeoRSCLIP  & & ViT-B/32      & 0.132         & 0.270        \\
GeoRSCLIP  & \cross & ViT-L/14      & 0.161         & 0.34         \\
 RemoteCLIP & & ViT-B/32      & 0.124         & 0.274        \\
 RemoteCLIP & & ViT-H/14      & 0.117         & \textbf{0.482}        \\ 
 DINOv2     & & ViT-L/14      & \textbf{0.306}         & 0.416        \\ \hline
 CLIP       & & ViT-B/32      & 0.190         & 0.098        \\
 CLIP       & & ViT-L/14      & 0.215         & 0.451        \\
 GeoRSCLIP  & & ViT-B/32      & 0.097         & 0.228        \\
GeoRSCLIP  & \checkmark & ViT-L/14      & 0.224         & 0.420        \\
 RemoteCLIP & & ViT-B/32      & 0.116         & 0.229        \\
 RemoteCLIP & & ViT-H/14      & 0.086         & \textbf{0.452}       \\ 
 DINOv2     & & ViT-L/14      & \textbf{0.358}         & 0.377        \\ \hline   
\end{tabular}
}
\caption{Performance (mAP) of different backbones for 10-shot on the SIMD dataset, including a general VLM (CLIP), remote sensing VLMs (GeoRSCLIP and RemoteCLIP) and a purely visual model (DINOv2). Prototypes without fine-tuning are shown on top, while fine-tuned prototypes are reported on the bottom. One negative example per image was used during training. Visual features show higher detection capabilities on novel classes $c_{novel}$, and they report strong performance in base classes $c_{base}$ as well. DINOv2 largely outperforms RemoteCLIP on novel classes despite having fewer parameters.}
\label{table:backbone_ablation}
\end{table}

\pparagraph{Classification abilities of DINOv2 features. } The results reported in Table~\ref{table:main_table_results} illustrate impressive detection performance on SIMD and a considerable decline when it comes to the DIOR dataset. We hypothesize this is mainly due to one aspect; several objects in DIOR significantly differ from the types of objects in DOTA, thus the pre-trained RPN provides unsuitable proposals for such categories. More precisely, DIOR contains some classes involving land cover areas and buildings, such as \textit{airport}, \textit{trainstation}, \textit{dam} or \textit{toll-station}. These substantially differ from the concept of \textit{object} in DOTA, which considers building and ground areas as background elements. Therefore, categories containing those elements in DIOR will be ignored as object candidates and consequently never detected. To clarify this hypothesis, we evaluate the classification abilities of the learned prototypes using their ground truth box annotations as region proposals. A strong classification performance and a decrease in the disparity between SIMD and DIOR results would indicate that the RPN is indeed a limiting factor of the approach for the DIOR dataset. Hence, we report in Table~\ref{table:classification_ablation} the classification F-1 score and accuracy of pre-trained prototypes using bounding boxes for the SIMD and DIOR datasets on 5-shot, 10-shot and 30-shot. Furthermore, to minimize background information in object prototypes, we additionally initialize and fine-tune them on the same subsets using segmentation masks instead of bounding boxes. Thus, a segmentation mask is extracted for each object using the Segment Anything Model (SAM)\cite{sam} by using its ground truth bounding box as an input prompt. Results indicate an impressive classification ability for prototypes learned via both boxes and segmentation masks. To complement this analysis, we select the best performing FSOD method on DIOR, FSRW~\cite{fsrw}, and use it as RPN to pair it with the prototypical classification. FSRW re-weights the features of a pre-trained one-stage detector and thus improves the bounding box proposals with respect to the base training. As reported in Table~\ref{table:main_table_results}, re-classifying FSRW proposals with our approach improves FSRW itself for the DIOR dataset. This ablation clarifies one aspect of our detector: selecting an RPN pre-training dataset with an object definition that better aligns with your target classes will considerably increase the performance. Alternatively, one can re-purpose FSRW as RPN if no suitable dataset is available, as shown. Lastly, the impact of fine-tuning using segmentation masks on classification remains unclear. Given the substantial computational overhead introduced by SAM, we opt to continue using exclusively bounding boxes.

\begin{table}[t]
\centering
\resizebox{\columnwidth}{!}{
\begin{tabular}{ccccccc}
\toprule
& \multirow{2}{*}{\shortstack{\textbf{fine-tuning}\\ \textbf{type}}} & \multicolumn{2}{c}{\textbf{SIMD}} & \multicolumn{2}{c}{\textbf{DIOR}} \\
\cmidrule(lr{0.125em}){3-4}
\cmidrule(lr{0.125em}){5-6}
& & \textbf{F-1 score} & \textbf{Acc} & \textbf{F-1 score} & \textbf{Acc} \\
\cmidrule(lr{0.125em}){3-3}
\cmidrule(lr{0.125em}){4-4}
\cmidrule(lr{0.125em}){5-5}
\cmidrule(lr{0.125em}){6-6}
\multirow{2}{*}{\textbf{5-shot}}        & boxes & \textbf{57.96} & \textbf{62.43} & \textbf{59.58} & \textbf{73.35} \\
                                        & masks & 50.49 & 58.17 & 48.98 & 62.68 \\
\midrule 
\multirow{2}{*}{\textbf{10-shot}}       & boxes & \textbf{64.30} & \textbf{69.23} & 60.60 & 75.74 \\
                                        & masks & 63.42 & 68.23 & \textbf{71.88} & \textbf{86.67} \\
\midrule
\multirow{2}{*}{\textbf{30-shot}}       & boxes & \textbf{66.88} & \textbf{68.95} & \textbf{72.23} & \textbf{84.36} \\
                                        & masks & 60.97 & 66.65 & 72.02 & 80.13 \\
\bottomrule
\end{tabular}
}
\caption{Classification results for fine-tuned prototypes using both boxes and masks (retrieved using SAM). The F-1 score and classification accuracy are provided for the SIMD and DIOR datasets, on 5, 10 and 30-shot, respectively. }
\label{table:classification_ablation}
\end{table}

\pparagraph{Impact of prototype fine-tuning. } We complement our evaluation with an ablation of the proposed prototype fine-tuning. Consequently, we report the mAP50 results for the SIMD dataset with $N=\{5,10,30\}$ without model fine-tuning, fine-tuning without negative examples, i.e. only fine-tuning object prototypes, and fine-tuning with 1, 5, and 10 negative examples per image, respectively. Table~\ref{table:fine_tuning_ablation} shows the obtained results. As seen, the best scores in all cases correspond to prototype fine-tuning with no negative examples. Interestingly, we use one negative example per image in Table~\ref{table:backbone_ablation} experiments as we observed that CLIP backbones yield a poor performance otherwise, assigning nearly all proposals to background prototypes. We attribute this behavior to the fact that captions in satellite images often describe the land cover, thus CLIP is more familiar with the background of objects than the objects themselves. Conveniently, DINOv2 does not require negative examples, enabling a considerable speed-up of the fine-tuning process as only a few prototypes are optimized. We believe this to be due to the way class-reference prototypes are learned. When using negative examples and training both object and background prototypes altogether, we are trying to yield vectors that separately characterize objects and the background. However, region proposals will partially contain the background in addition to the object. Furthermore, DINOv2 representations not only describe local information but also the relationship with nearby patches, i.e. the background. Consequently, learning DINOv2 reference vectors that completely decouple foreground from background information might be ill-posed.

\begin{table}[t]
\centering

\begin{tabular}{ccccc}
\hline
\textbf{fine-tuning} & \textbf{\# of negatives} & \textbf{N=5} & \textbf{N=10} & \textbf{N=30} \\ \hline
\cross               & 0 & 0.297 & 0.303 & 0.339 \\
\checkmark           & 0 & \textbf{0.354} & \textbf{0.390} & \textbf{0.412} \\
\checkmark           & 1 & 0.336 & 0.362 & 0.382 \\
\checkmark           & 5 & 0.322 & 0.363 & 0.349 \\
\checkmark           & 10 & 0.308 & 0.365 & 0.333 \\ \hline
\end{tabular}

\caption{Ablation of the proposed prototype fine-tuning approach. Different numbers of examples per class $N=\{5, 10, 30\}$ are evaluated on mAP50 for prototypes with no fine-tuning, fine-tuning only object prototypes, and fine-tuning with 1, 5 and 10 negative boxes per image, respectively. }
\label{table:fine_tuning_ablation}
\end{table}

\subsection{Limitations and future work.}
\pparagraph{Region proposals.} Our framework can be used for two different applications: On one hand, one could seek fine-grained detection of common, general objects, e.g. vehicles or aircraft, as is the case of the SIMD dataset. This scenario is very suitable to our approach, as the RPN is capable of reliably detecting the general object, while the learned prototypes allow for fine-grained classification of its sub-categories. On the other hand, one could want to detect new, unseen categories that significantly differ from the ones in the pre-training dataset. As illustrated by the DIOR results, the RPN limits the performance of the method, as it fails to provide suitable region proposals for these challenging objects. Even though using FSRW as RPN improves the proposed bounding boxes, future work should focus on a better adaptation of RPN to novel classes using available annotations and prototypes.

\pparagraph{Classifying boxes.} Features inside region proposals contain object and background information. Hence, using all patch similarities inside the box for classification might negatively impact the results. In our experiments, averaging all similarities yielded better results than taking the maximum similarity or the top $k$ most similar patches. Nevertheless, this could be studied in depth in further analysis.

\section{Conclusion}
In this article, we thoroughly explore recent ideas in open-vocabulary and few-shot object detection for remote sensing applications. More precisely, we develop a few-shot object detector that builds prototypes of objects and their backgrounds using robust features and fine-tunes them via automatic prompt engineering. Furthermore, we explore the use of visual (DINOv2) and vision-language (CLIP) representations, including two VLMs specifically tailored for the remote sensing domain. We find that visual features are largely superior to vision-language representations, as they do not have the vocabulary and/or knowledge for fine-grained remote sensing object detection. We demonstrate the capabilities of DINOv2 features to represent and classify rare objects in satellite imagery with only a handful of examples. Lastly, we compare our simple approach with other fully supervised and few-shot methods on two challenging datasets, SIMD and DIOR, for 5,10, and 30-shot. Our approach provides large improvements for the SIMD dataset, while a simple change of region proposal network allows us to beat other methods on the DIOR dataset as well.

%
{
    \small
    \bibliographystyle{ieeenat_fullname}
    \bibliography{main}
}

\clearpage
\setcounter{page}{1}
\maketitlesupplementary

\appendix
\section{Datasets}
\label{sec:dataset_details}
As mentioned in the main text, satellite images are commonly imbalanced and contain several objects in a single image. This hinders the process of randomly selecting a subset with an equal number of instances per category. Hence, some classes of our subsets have a few examples more or less than $N$. We report the exact number of instances per category in Table~\ref{table:simd_num_instances} and Table~\ref{table:DIOR_num_instances}, for datasets SIMD and DIOR, respectively. In addition, we clarify in the tables which categories belong to novel classes and which ones are base classes. The class \textit{others} of the SIMD dataset is removed from all evaluations, as it is highly underrepresented in the dataset and selecting a subset of approximately $N$ samples per class which includes the \textit{others} category is not trivial. The data splits will be publicly released, containing the images and annotations of each of the subsets.





\begin{table}[t]
\centering
\begin{tabular}{lcccc} \hline
\textbf{Type} & \textbf{Category}          & \textbf{N=5} &\textbf{ N=10} & \textbf{N=30} \\ \hline
\multirow{4}{*}{\textbf{c}$_{base}$}  & \textit{car}                  & 5   & 11   & 32   \\
                    & \textit{helicopter}          & 5   & 10   & 30   \\
                    & \textit{long-vehicle}        & 5   & 9    & 29   \\
                    & \textit{boat}               & 6   & 10   & 33   \\ \hline
\multirow{10}{*}{\textbf{c}$_{novel}$} & \textit{truck}                & 5   & 10   & 30   \\
                    & \textit{van}                   & 5   & 11   & 33   \\
                    & \textit{bus}                  & 5   & 10   & 31   \\
                    & \textit{airliner}             & 8   & 11   & 36   \\
                    & \textit{propeller-aircraft}   & 5   & 10   & 28   \\
                    & \textit{trainer-aircraft}     & 6   & 10   & 30   \\
                    & \textit{charted-aircraft}     & 5   & 10   & 30   \\
                    & \textit{fighter-aircraft}    & 5   & 11   & 30   \\
                    & \textit{stair-truck}          & 5   & 10   & 34   \\
                    & \textit{pushback-truck}       & 5   & 10   & 30   \\ \hline
\end{tabular}
\caption{Number of instances per class $N$ for each of the used subsets of the SIMD dataset, i.e. $N=\{5, 10, 30\}$. Classes are divided between base classes $\textbf{c}_{base}$ and novel classes $\textbf{c}_{novel}$.}
\label{table:simd_num_instances}
\end{table}


\begin{table}[t]
\centering
\begin{tabular}{lcccc} \hline
\textbf{Type} & \textbf{Category}          & \textbf{N=5} &\textbf{ N=10} & \textbf{N=30} \\ \hline
\multirow{8}{*}{\textbf{c}$_{base}$}  & \textit{ship}                 & 5   & 10   & 33   \\
                    & \textit{harbor}               & 5   & 11   & 32   \\
                    & \textit{baseballfield}        & 5   & 10   & 30   \\
                    & \textit{groundtrackfield}     & 5   & 10   & 30   \\
                    & \textit{tenniscourt}          & 5   & 10   & 31   \\
                    & \textit{storagetank}          & 5   & 10   & 32   \\
                    & \textit{airplane}             & 5   & 10   & 30   \\
                    & \textit{basketballcourt}      & 5   & 10   & 31   \\ \hline
\multirow{12}{*}{\textbf{c}$_{novel}$} & \textit{chimney}              & 5   & 10   & 30   \\
                    & \textit{vehicle}              & 5   & 11   & 31   \\
                    & \textit{airport}              & 5   & 10   & 30   \\
                    & \textit{golffield}            & 5   & 10   & 30   \\
                    & \textit{overpass}             & 5   & 10   & 30   \\
                    & \textit{bridge}               & 5   & 10   & 30   \\
                    & \textit{express-toll-station} & 5   & 10   & 30   \\
                    & \textit{stadium}              & 5   & 10   & 30   \\
                    & \textit{trainstation}         & 5   & 10   & 30   \\
                    & \textit{express-service-area} & 5   & 10   & 30   \\
                    & \textit{windmill}             & 5   & 10   & 31   \\
                    & \textit{dam}                  & 5   & 10   & 30   \\ \hline
\end{tabular}
\caption{Number of instances per class $N$ for each of the used subsets of the DIOR dataset, i.e. $N=\{5, 10, 30\}$. Classes are divided between base classes $\textbf{c}_{base}$ and novel classes $\textbf{c}_{novel}$.}
\label{table:DIOR_num_instances}
\end{table}

\begin{figure*}
    \centering
    
    \begin{subfigure}[t]{0.49\linewidth}
        \centering
        \includegraphics[width=\linewidth,clip]{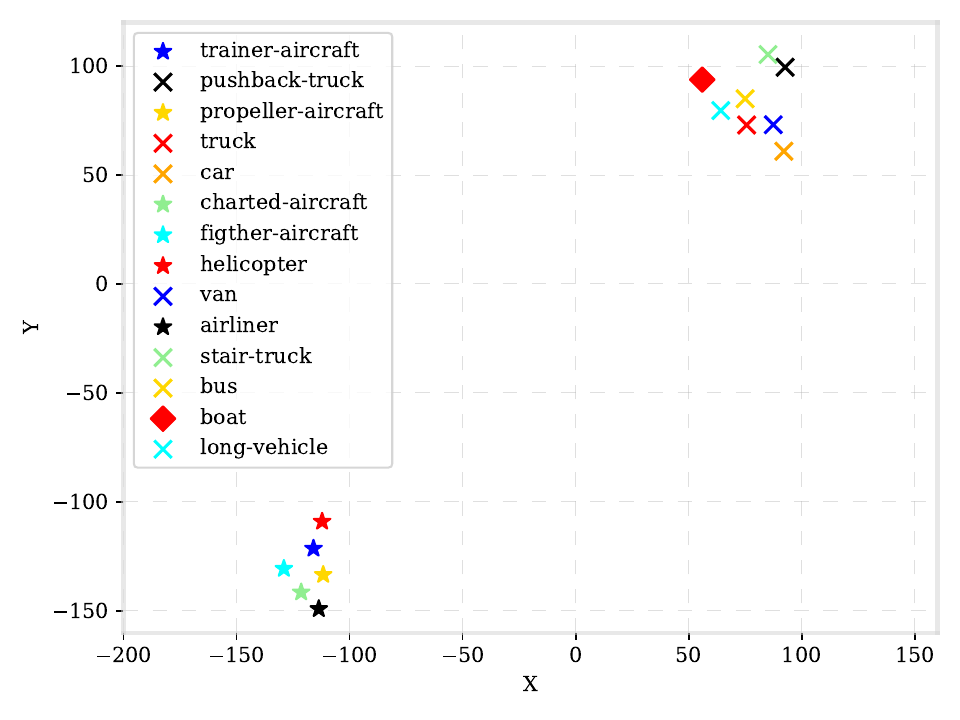}
        \caption{Prototypes without fine-tuning}
        \label{fig:sub2}
    \end{subfigure}
    \hfill
    \begin{subfigure}[t]{0.49\linewidth}
        \centering
        \includegraphics[width=\linewidth,clip]{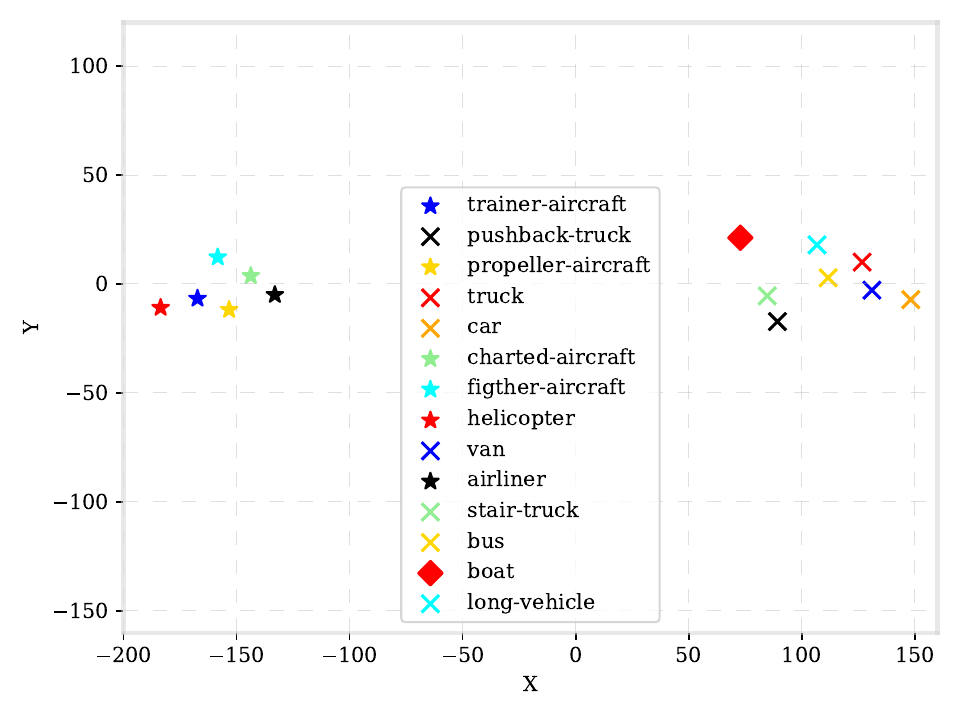}
        \caption{Prototypes after fine-tuning}
        \label{fig:sub1}
    \end{subfigure}
    \caption{T-SNE visualization of the learned prototypes for the SIMD dataset using $N=10$, before and after fine-tuning. Plane or aircraft types are shown with a star marker, while types of terrestrial vehicles are shown with a cross marker. The \textit{boat} class is shown as a diamond. 
    As depicted, class separation increases after fine-tuning, e.g. \textit{stair-truck} and \textit{pushback-truck} are more separable after training. In addition, each cluster representing a group of transportation exhibits close proximity yet remains distinguishable, whereas the separation between other groups is more pronounced.}
    \label{fig:simd_tsne}
\end{figure*}
\section{Implementation details}
In this section, we provide further implementation details concerning our evaluation.

\pparagraph{Ours.} We train our model using Adam optimizer~\cite{adam} over $200$ epochs and a learning rate of $2e^{-4}$. We reduce the learning rate by a factor of $0.1$ at epochs 10 and 100. As mentioned in the main text, we apply spatial and radiometric transformations, which involve horizontal and vertical flips with $0.5$ probability each, random 90-degree rotation with $0.5$ probability, color jitter with $brightness=0.2$, $contrast=0.2$, $saturation=0.2$, and $hue=0.1$, padding with $0.5$ probability, and random crops at a scale of $0.5$ to $1$. Lastly, crops are resized to $602\times 602$.

\pparagraph{YOLO.} We use the source code of YOLOv5 by Ultralytics to pre-train a \textit{yolov5s} model on the entire DOTA dataset. We use 200 epochs with a batch size of 64, using the Ultralytics pre-defined hyper-parameters. The best model was kept and used from there on. Subsequently, we fine-tuned the learned model on the different subsets in two ways. First, we re-train the model on each few-shot subset over 200 epochs with a batch size of 128. Then, we repeat the process but freeze the model's backbone and fine-tune only the heads, thus avoiding overfitting the pre-trained image representations on the small amount of available data. In both setups, we report the results of the best models after the few-shot training.

\pparagraph{DE-ViT.} We use the official implementation of DE-ViT as by their authors. We create masks using the annotations of each subset and generate prototypes with their code. Subsequently, we use their pre-trained model for evaluation of the test datasets with the default values suggested in their publication.

\pparagraph{FSRW.} We use the official implementation of the FSRW as by their authors. We perform the model's full training on our end, i.e. training the model in original data and few-shot fine-tuning using the SIMD and DIOR subsets. The hyper-parameters are kept by default as provided by the authors.

\section{Visualization of learned prototypes}
Figure~\ref{fig:simd_tsne} provides the T-SNE visualization of the learned prototypes before and after fine-tuning, for the SIMD dataset. We can observe a large separation between groups of classes that represent different types of transportation, i.e. types of planes or aircraft, types of vehicles, and \textit{boat}. Furthermore, the comparison shows that class separability increases after prototype fine-tuning. For example, the class \textit{boat} is quite close to \textit{long-vehicle} and \textit{bus} before fine-tuning, and the distance increases afterward. Similarly, the separability of \textit{stair-truck} and \textit{pushback-truck} increases after learning the prototypes.


\end{document}